\documentclass[10pt,twocolumn,letterpaper]{article}

\usepackage[pagenumbers]{cvpr} 

\usepackage{enumitem}
\usepackage[most]{tcolorbox}
\usepackage{minted}

\usepackage{multirow}
\usepackage{bbding}

\usepackage[table,xcdraw]{xcolor}

\newcommand{\selfcall}{\textit{think-through-self-calling}}
\newcommand{\subagent}{\textrm{SubagentVL}}
\newcommand{\vstar}{\textit{V*}}
\newcommand{\modelname}{sCoT}

\definecolor{cvprblue}{rgb}{0.21,0.49,0.74}
\usepackage[pagebackref,breaklinks,colorlinks,allcolors=cvprblue]{hyperref}

\def\paperID{13599} 
\def\confName{CVPR}
\def\confYear{2026}

\title{Thinking with Images via Self-Calling Agent}
\author{Wenxi Yang\textsuperscript{1}\footnotemark[1] \quad
Yuzhong Zhao\textsuperscript{1}\footnotemark[1] \quad
Fang Wan\textsuperscript{1} \quad
Qixiang Ye\textsuperscript{1$\dagger$}
\\
\textsuperscript{1}University of Chinese Academy of Sciences\\
{\tt\small yangwenxi24@mails.ucas.ac.cn} \quad 
{\tt\small zhaoyuzhong20@mails.ucas.ac.cn} \quad \\
{\tt\small wanfang@ucas.ac.cn} \quad
{\tt\small qxye@ucas.ac.cn}
}

\begin{document}
\twocolumn[{
\renewcommand\twocolumn[1][]{#1}%
\maketitle

\begin{center}
    \centering
    \captionsetup{type=figure}
    \includegraphics[width=0.68\textwidth]{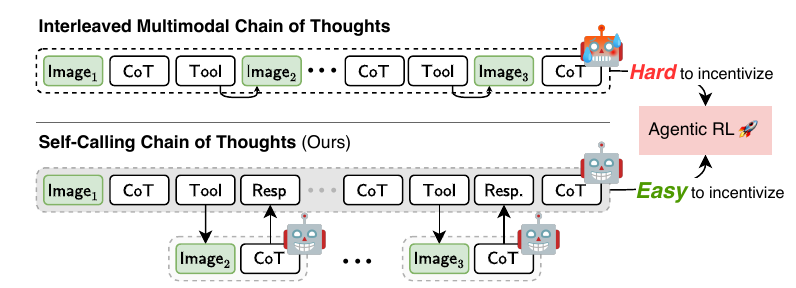}
    \includegraphics[width=0.27\textwidth]{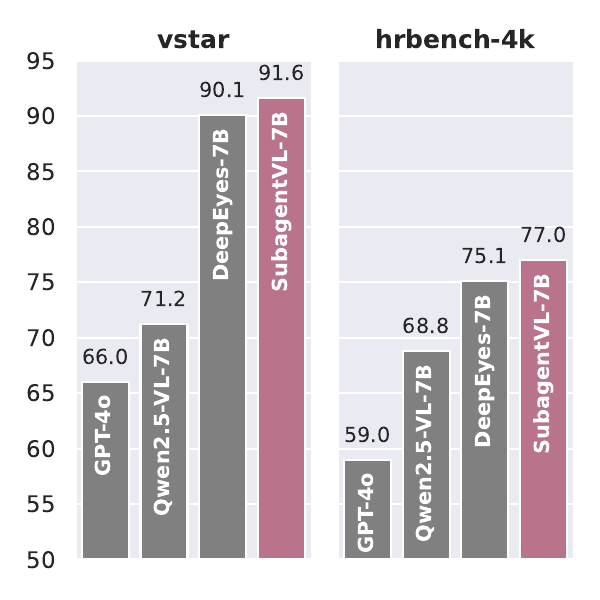}
    
    \vspace{1em}
    \captionof{figure}{\textbf{Comparison of interleaved multimodal and Self-Calling Chain-of-Thought (sCoT).} 
    \textit{Left:} The interleaved multimodal chain of thought (iMCoT) in DeepEyes~\citep{zheng2025deepeyes} requires agents to perform tightly interleaved vision–language reasoning, which is difficult to incentivize. In contrast, the proposed sCoT reformulates the process as a language-only reasoning trajectory (the context may contains images, but the reasoning itself is language-only), making multimodal reasoning much easier to incentivize.
    \textit{Right:} The model trained with sCoT outperforms interleaved multimodal chain-of-thoughts upon the challenging benchmarks {with $\sim75\%$ fewer GPU hours.}
    }
    
	\label{fig:comp}
\end{center}

}]

\renewcommand{\thefootnote}{\fnsymbol{footnote}}
\footnotetext[1]{~Equal contribution. $\dagger$ Corresponding Author.}

\maketitle
\begin{abstract}

\noindent Thinking-with-images paradigms have showcased remarkable visual reasoning capability by {integrating visual information as dynamic elements into the Chain-of-Thought (CoT).} 
However, optimizing interleaved multimodal CoT (iMCoT) through reinforcement learning remains challenging, as it relies on scarce high-quality reasoning data.
In this study, we propose \textbf{Self-Calling Chain-of-Thought} (\modelname{}), a novel visual reasoning paradigm that reformulates iMCoT as a 
language-only CoT
with \textbf{self-calling}. 
Specifically, a main agent decomposes the complex visual reasoning task to atomic subtasks and invokes its virtual replicas, \ie parameter-sharing subagents, to solve them in isolated context. 
%
%
\modelname{} enjoys substantial training effectiveness and efficiency, as it requires no explicit interleaving between modalities.
\modelname{} employs group-relative policy optimization to reinforce effective reasoning behavior to enhance optimization. 
Experiments on HR-Bench 4K show that \modelname{} {improves the overall reasoning performance by up to $1.9\%$ with $\sim75\%$ fewer GPU hours compared to strong baseline approaches.
%
Code is available at this
{\href{https://github.com/YWenxi/think-with-images-through-self-calling.git}{url}}.
}


\end{abstract}    
\section{Introduction}

\begin{figure*}
    \centering
    \includegraphics[width=\linewidth]{sec/figs/chat.pdf}
    \caption{\textbf{Thinking-with-images-through-self-calling.}
A visual language model termed \textit{main agent} decomposes a complex visual query into simple, atomic subtasks handled by the model's virtual replicas (``\textit{subagents}''). Each subagent solves a localized atomic subtask (grounding, captioning, OCR, \etc) and returns textual outputs which are aggregated by the main agent to produce the final answer.}
    \label{fig:chat}
\end{figure*}

\label{sec:intro}
The thinking-with-images paradigms like OpenAI o3~\citep{wu2024v, wu2024mind, zheng2025deepeyes, qiao2025vthinker, su2025thinking, su2025openthinkimg, xu2025visualplanning}, which are built on powerful interleaved multimodal large language models (MLLMs), have demonstrated impressive visual reasoning capabilities by incorporating images as dynamic elements in the reasoning process. 
A key factor behind their success is reinforcement learning, which incentivizes valuable interleaved multimodal chains-of-thought (iMCoT) from MLLMs and further stimulates the emergence of ``thinking with images'', Fig.~\ref{fig:comp}(top left).
%

However, incentivizing iMCoT in MLLMs through reinforcement learning is substantially more challenging than training non-interleaved or language-only Chain-of-Thought (CoT), {as existing interleaved multimodal reasoning corpora are relatively scarce compared to language-only reasoning data}.
This limitation not only makes iMCoT agents hard to optimize, but also undermines their reasoning and tool-calling capabilities.

In this study, we propose the \textit{\textbf{Self-Calling Chain-of-Thought}} (\modelname{}), a novel visual reasoning paradigm that reformulates iMCoT as language-only CoT augmented with self-calling, \cref{fig:comp}(bottom left).
%
A vision-language model, termed the \textit{main agent}, decomposes a complex visual reasoning task into a sequence of simple atomic subtasks, termed subagents, identical to the language model itself. 
These atomic subtasks are sufficiently simple to be solved through language-only CoT.
In particular, subagents are designed as virtual replicas, which shares exactly the same weights with the main agent.
Once all the subtasks are accomplished by the subagents, the main agent aggregates their textual responses and derives the final answer.

%
The proposed sCoT can be effectively incentivized through end-to-end reinforcement learning. 
We optimize the main agent’s reasoning trajectory using group-relative policy optimization~\cite{shao2024deepseekmath} (GRPO), while masking out the textual outputs returned by subagents to avoid reward leakage.
This setup allows the model to learn effective self-calling behaviors—when to call subagents and how to structure subtasks—without directly overfitting to subagent responses.
%
%
With above design, the proposed \modelname{} paradigm does not require interleaved multimodal reasoning capabilities from the agents, \cref{fig:comp}(bottom left), which makes it much easier to optimize than iMCoT.
This design also facilitates saving memory and enhancing generalization of the reasoning capabilities learned by the agents, \cref{fig:chat}.

Experiments conducted on $\vstar$ benchmark and HR-Bench 4K demonstrate that \modelname{} delivers superior visual reasoning capabilities.
As shown in \cref{fig:comp}(right), it outperforms the state-of-the-art iMCoT approach DeepEyes~\citep{zheng2025deepeyes} by 1.2\% and 1.9\% on $\vstar$ benchmark and HR-Bench 4K respectively.
In addition, since \modelname{} is easier to incentivize than iMCoT, it incurs much lower training costs compared to DeepEyes, thich is trained with iMCoT.
Specifically, \modelname{} achieves better results than Deepeyes with only $\sim$25\% of the GPU hours. 
\section{Related Work}
\label{sec:related-work}

\paragraph{Thinking with Images} extends Chain-of-Thought (CoT) reasoning from the language area to the multimodal domain by enabling MLLMs to actively inspect visual content~\cite{o3}. Prior studies enhance perception through exploration tools such as zooming, object detection, and segmentation~\cite{yang2023set, ma2024taco, qi2025cogcom, wu2024mind, wang2025visuothink, hu2022promptcap},
allowing models to examine localized regions of an image.
Other approaches strengthen visual reasoning by designing multimodal CoTs —$e.g.$, Visualization-of-Thought~\cite{wu2024mind} and Visual Abstract Thinking~\cite{liu2025visual}—or by constructing manually defined workflows and tool chains that convert visual information into language-friendly intermediate representations~\cite{cheng2025visual, wu2024dettoolchain}.
Unlike these pipelines, this study bridges the modality gap without external experts or handcrafted workflows. 
We instead leverage the MLLM’s intrinsic visual abilities through self-invoked atomic subtasks and use language\textemdash only reasoning to coordinate them, yielding a lightweight and flexible alternative to existing multimodal CoT systems.

\paragraph{Reinforcement Learning.}
Recent advances in reinforce-
ment learning, such as GRPO~\cite{shao2024deepseekmath} and its variants~\cite{zhao2025geometric,yu2025dapo,yue2025vapo,zheng2025group,liu2025understanding,chen2025seed} offer powerful paradigms for optimizing policies in sequential decision-making.
These approaches naturally extend to multimodal large language models (MLLMs)~\cite{bai2025qwen25vltechnicalreport,cheng2025visual,huang2025vision,zhao2025r1,shen2025vlm,chen2025grpo} 
and agentic frameworks~\cite{feng2025retoolreinforcementlearningstrategic,dong2025aepo,dong2025tool,dong2025arpo,huang2025vision}, 
where RL is used to learn when and how to perceive, reason, and act across visual and textual modalities. 
In thinking-with-images settings, prior work (e.g., Chain-of-Focus~\cite{zhang2025chain}, ACTIVE-o3~\cite{zhu2025active}, DeepEyes~\cite{zheng2025deepeyes}) typically relies on multimodal CoTs with interleaved visual observations and tool invocations to guide region selection and tool usage. 
Nevertheless, such heterogeneous reasoning trajectories demand consistent reasoning across modalities, making them difficult to incentivize directly through reinforcement learning. 
In contrast, we adopt a pure language-based CoT, applying reinforcement learning only to textual reasoning while accessing tools via self-calling.

\begin{figure*}[t]
    \centering
    \includegraphics[width=0.7\linewidth]{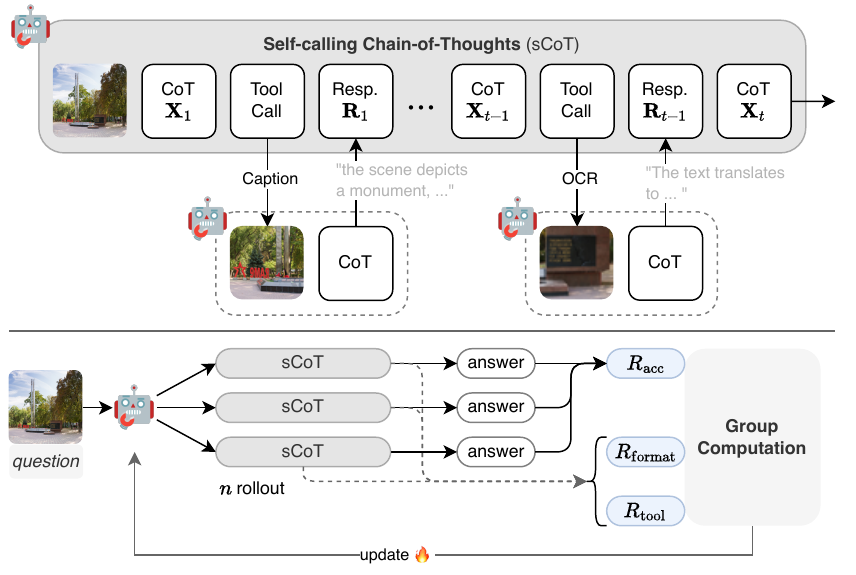}
    \caption{\textbf{Diagram of sCoT.} 
    During training, the main agent receives a task given by users and decomposes it into subtasks and invokes parameter-sharing subagents iteratively through pure-language CoT steps. This self-calling mechanism allows the model to simulate tool-using behaviors purely in the language domain. Then, the resulting CoT responses are used to calculate rewards, which includes three aspects: accuracy (\(R_{\text{acc}}\)), format (\(R_{\text{format}}\)), and tool-usage (\(R_{\text{tool}}\)). Finally, GRPO is used to incentivize \modelname{}.}
    \label{fig:diagram}
\end{figure*}

\section{Preliminary: Thinking with Images}
\label{sec:preliminary}

The reasoning process of thinking-with-images can be formulated as an interleaved Multimodal Chain-of-Thought (iMCoT), in which the reasoning modality alternates between textual and visual domains~(\cref{fig:comp}, top left). 

While iMCoT encourages fine-grained visual reasoning, 
it produces long interleaved chains in which images and text alternate, 
placing heavy reasoning burden on current MLLMs. 
Empirically, state-of-the-art models match specialized detectors on single-image tasks suffer pronounced performance degradation as the number of images increases~\cite{wu2025visual}.
This degradation becomes more severe when the model must integrate cues across multiple key images or under noisy conditions.
For instance, open-source models such as Qwen2-VL-7B-Instruct suffer more than 15\% accuracy drops in two-image visual needle-in-a-stack benchmark~\cite{wu2025visual}, highlighting the difficulty of scaling visual reasoning beyond single-image contexts.

\section{The Proposed Approach}

\label{sec:method}

\subsection{Self-Calling Chain-of-Thought}
State-of-the-art MLLMs already excel at simple, atomic vision–language tasks such as grounding, OCR, and captioning. 
Motivated by this, we propose the \textit{\textbf{Self-Calling Chain-of-Thought}}~(\modelname{}), an alternative reasoning formulation that avoids jointly processing interleaved modalities. 
As shown in \cref{fig:chat}, a complex visual language task are decomposed to a sequence of atomic subtasks, each aligned with fundamental visual capabilities that the MLLM excels. 
To reduce the reasoning burden of each step, each subtask is processed in an isolated context, independent of the main reasoning chain.
The main reasoning chain then aggregates the textual responses returned by these subtasks to derive the final answer.

\paragraph{Pipeline.} The overall pipeline of sCoT is illustrated in \cref{fig:diagram}. 
Three elements play important roles.

\begin{itemize}
    \item \textbf{Main Agent:} We call the MLLM which receives the user question as the main agent, which understands the user question and decompose it into atomic subtasks. After all subtasks are accomplished, the main agent collects all the textual responses and aggregates them to a final answer, Figs.~\ref{fig:chat} and ~\ref{fig:diagram}.
    
    \item \textbf{Subtask:} A subtask consists of 
    \begin{enumerate}[(a)]
        \item \textit{a simple yet clear question},
        \item \textit{a cropped subregion of the image in the original task}
    \end{enumerate}
    
    \item \textbf{Subagent:} Each subagent is an identical \textit{virtual replica} of the main agent. 
\end{itemize}
Here, a virtual replica means that each subagent is a unmaterialised instance of the same model--sharing the exact weights and parameters--without deploying any additional physical models. 
In practice, invoking/calling a subagent simply issues another forward call to the LLM server that runs the main agent. 
Each subagent focuses on a single isolated subtask, processes one cropped image at a time, and returns a concise textual response to the main agent.

\paragraph{Tool Calling Protocol.}
\label{sec:tool-calling-protocol}
The self-calling interaction is realized through structured tool calls, enabling the main agent to invoke subagents dynamically during reasoning. 
Each invocation is parameterized by three arguments, as
\begin{enumerate}[\indent (1)]
    \item a \textit{task type} \texttt{(str)}, which briefly specifies the subtask category (\eg, OCR, visual question answering, caption, \etc); 
    \item a \textit{prompt} \texttt{(str)}, representing the textual query or instruction for the subtask; and 
    \item a \textit{bounding box} \texttt{(list[int])}, defining the spatial region of interest within the image. 
\end{enumerate}
Subagents operate independently across tool calls, without access to global information or communication with other agents, ensuring isolated and modular reasoning.
To provide limited local context while maintaining focus, and inspired by region-level understanding methods such as ControlCap~\cite{zhao2024controlcap} and DynRefer~\cite{zhao2025dynrefer}, each bounding box is slightly enlarged through the following Equation
\begin{equation}
    \text{Bbox} = \text{Bbox}_{\text{ground}} \cdot (1-\alpha) + \text{Bbox}_{\text{canvas}}\cdot \alpha,
\end{equation}
where $\alpha$ is the interpolation factor (0.05 by default). $\text{Bbox}_{\text{ground}}$ and $\text{Bbox}_{\text{canvas}}$ are the bounding boxes of the grounding result and the size of the image canvas respectively. The enlargement allows the subagents to capture additional contextual information for more comprehensive region-level understanding.

\begin{tcolorbox}[colback=white,colframe=purple!70!gray!70!,title=\textbf{Subtask Prompt Template}]
\textbf{System:} You are a helpful assistant serving as a subagent. Given a cropped image and a question, answer the question and return the result. If further information is needed, encourage the user to make another call to the tool.

\textbf{User:}
\texttt{\{cropped image\}} 

[subtask: \texttt{\{task\_type\}}] \texttt{\{prompt\}}
\end{tcolorbox}

\subsection{Agentic Reinforcement Learning}




%
The next challenge is enabling the model to learn orchestration of self-calling interactions. 
Rather than manually defining workflows in the system prompt or relying on in-context examples, we employ end-to-end agentic reinforcement learning to explicitly reward coherent self-reflection and efficient subtask coordination.

\paragraph{Rollout.}
Although our reasoning traces resemble language-only CoT, the underlying decision process is still naturally expressed as a Markov decision process composed of interleaved states and responses. At step $t$, the state $s_t$ of CoT is defined as
\begin{equation}
    s_t = \{ (\mathbf X_1, \mathbf R_1), (\mathbf X_2, \mathbf R_2), \dots, (\mathbf X_t, \mathbf R_t) \},
\end{equation}
where \( \mathbf R_{\leq t} = \{\mathbf R_1, \mathbf R_2,..., \mathbf R_t\}\) denotes the accumulated \textbf{textual responses} from subagents (subtask results), and \( \mathbf X_{\leq t} = \{\mathbf X_1, ..., \mathbf X_t\} \) represents the reasoning tokens generated by the main agent up to step \(t\).
These textual tokens may include intermediate thoughts, subtask plans, and summary statements.
Therefore, we name $s_t$ as the \textit{\textbf{Self-Calling Chain-of-Thought}} (\modelname{}). 
%
The agent now learns a policy that maximizes long-term reward over this language-only reasoning trajectory, effectively aligning self-reasoning with the goal of accurate and efficient visual understanding.

\paragraph{Reward Design.}
We adopt the reward design from DeepEyes~\cite{zheng2025deepeyes} with a minor modification to prevent potential reward hacking. For a rollout trajectory $\tau$, the total reward is defined as
\begin{equation}
    \label{eq:reward}
    R(\tau) = R_{\rm acc} + R_{\rm format} + \mathbb{I}_{R_{\rm acc} > 0} \cdot {\mathbb{I}_{\rm tool \prec ans} }
    \cdot R_{\rm tool}
\end{equation}
where $R_{\text{acc}}$ measures the correctness of the final answer, and $R_{\text{format}}$ penalizes ill-structured or malformed outputs.
The key modification lies in the tool usage bonus $R_{\text{tool}}$, which originally is granted only when the model produces a correct answer and invokes at least one external perception tool during reasoning. 
Take a further step, to ensure causal consistency, the tool invocation must occur before the final answer is generated, and therefore an indicator $\mathbb{I}_{\rm tool \prec ans}$ is incorporated into Eq.~\eqref{eq:reward}.
This constraint encourages meaningful visual grounding while discouraging reward exploitation through post-hoc or redundant tool calls.

\paragraph{Optimization.} 
We employ Group Relative Policy Optimization (GRPO)~\cite{shao2024deepseekmath} as our reinforcement learning algorithm,
owing to its proven efficiency and stability across diverse reasoning and decision-making tasks. 
During training, all tool calls within a rollout are executed sequentially to preserve causal consistency between reasoning and perception steps. 
For multi-turn agent trajectories, we apply a token-wise loss mask to exclude gradients on observation tokens (\ie subagents' responses) that are not generated by the model itself, ensuring that optimization focuses solely on the model’s own reasoning and action outputs, although the main agent and subagents are running on the same weights.

\section{Experiment}


\subsection{Setting}

\paragraph{Baselines.}
To evaluate the effectiveness of DeepEyes, we assess the trained model, denoted as \subagent, on two comprehensive benchmarks: \textit{V*} \cite{wu2024v} and HR-Bench. Both benchmarks systematically measure the perception capability of MLLMs across diverse tasks—including OCR, visual prompting, and attribute recognition—using high-resolution images up to 8K. We compare \subagent \ against DeepEyes, which is trained via the same end-to-end reinforcement learning process. In addition, we include comparisons with state-of-the-art models (e.g., GPT-4o, o3) and specialized visual reasoning workflows such as SEAL, DyFo, and ZoomEye, to comprehensively benchmark the performance of our approach under varied reasoning paradigms.

\paragraph{Data.}
We train Qwen2.5-VL-7B-Instruct on the dataset collected by DeepEyes, which integrates diverse visual reasoning sources to enhance both perception and analytical capability. The corpus consists of three major subsets:

\begin{enumerate}[\indent (1)]
    \item \textit{Fine-grained data} (47\%), derived from the $V^*$ training set \cite{wu2024v}, containing high-resolution images and detailed perception questions designed to maximize the effectiveness of tool-based reasoning;
    \item \textit{Chart data} (30\%)~\cite{masry-etal-2022-chartqa}, composed of synthetic charts and graph images that enrich the diversity of visual elements and quantitative patterns; and
    \item \textit{Reasoning data} (23\%), incorporated from the ThinkLite-VL dataset~\cite{wang2025sota}, which broadens task diversity and strengthens the model’s higher-level analytical reasoning skills.
\end{enumerate}

\paragraph{Training Details.}
Comparing with DeepEyes, we adopt a similar yet simplified training setup for our Agentic RL process, as detailed in \cref{tab:training-config}. The training is conducted over the veRL framework. During training, the main agent invokes subagents in a sequential manner, \ie, subtasks are executed one after another rather than in parallel. For reward computation, the accuracy reward $R_{\mathrm{acc}} \in \{0, 0.8\}$ is assigned by an LLM-as-a-judge, while the format reward $R_{\mathrm{format}} \in \{-0.2, 0\}$ and the tool-usage reward $R_{\mathrm{tool}} \in \{0, 1.2\}$ are determined based on the correctness of output format and the existence and ordering of tool calls, respectively.

\begin{table}
\small
\caption{RL Training Configuration.}
\centering
\begin{tabular}{@{}lcc@{}}
\toprule
 & \begin{tabular}[c]{@{}c@{}}DeepEyes\\ \textit{\small think-with-images}\end{tabular} & \begin{tabular}[c]{@{}c@{}}\subagent\\ \textit{\small think-through-self-calling}\end{tabular} \\
\midrule
GPU & 8/32 H100 & 8 A100 \\
base model & Qwen2.5-VL-7B & Qwen2.5-VL-7B \\
batch size & 256 & 56 \\
rollout & 16 & 8 \\
LLM-judge & Qwen2.5-VL-72B & Qwen2.5-VL-7B \\
iterations & 80 & 80 \\
\bottomrule
\end{tabular}
\label{tab:training-config}
\end{table}

\subsection{Result}
\begin{table*}
\caption{Comparison on High-Rosolution Benchmarks. The mark * indicates reproduced results and $\dagger$ means results are evaluated using a Qwen2.5-7B-Instruct serving as a judge, while DeepEyes uses Qwen2.5-72B-Instruct.}
\begin{tabular}{lc|ccc|ccc|ccc}
\toprule
\textbf{} & \textbf{} & \multicolumn{2}{l}{\bf $V^*$ Benchmark} & & \multicolumn{2}{l}{\bf HR-Bench 4K} & & \multicolumn{3}{l}{\bf HR-Bench 8K}  \\
Model & Size & direct & relative & overall & FSP & FCP & Overall & FSP & FCP & Overall \\
\midrule
GPT-4o~\cite{o3} & - & - & - & 66.0 & 70.0 & 48.0 & 59.0 & 62.0 & 49.0 & 55.5 \\
 o3~\cite{gpt4o} & - & - & - & 95.7 & - & - & - & - & - & - \\
\midrule
\multicolumn{11}{l}{\it manually-defined-workflow} \\
SEAL~\cite{wu2024v} & 7B & 74.8 & 76.3 & 75.4 & - & - & - & - & - & - \\
DyFo~\cite{li2025dyfo} & 7B & 80.0 & 82.9 & 81.2 & - & - & - & - & - & - \\
ZoomEye~\cite{shen2025zoomeye} & 7B & 93.9 & 85.5 & 90.6 & 84.3 & 55.0 & 69.6 & 85.5 & 50.0 & 69.3 \\
\midrule
\multicolumn{11}{l}{\it baseline} \\
Qwen2.5-VL~\cite{bai2025qwen25vltechnicalreport} & 7B & 73.9 & 67.1 & 71.2 & 85.2 & 52.2 & 68.8 & 78.8 & 51.8 & 65.3 \\
Qwen2.5-VL~\cite{bai2025qwen25vltechnicalreport} & 32B & 87.8 & 88.1 & 87.9 & 89.8 & 58.0 & 73.9 & 84.5 & 56.3 & 70.4 \\
\midrule
\multicolumn{11}{l}{\it think-with-images} \\
DeepEyes${}^{*\dagger}$ & 7B & 91.3 & 82.9 & 88.0 & 92.0 & 58.3 & 75.1 & 85.0 & 57.0 & 71.0 \\
DeepEyes~\cite{zheng2025deepeyes} & 7B & 91.3 & 88.2 & 90.1 & 91.3 & 59.0 & 75.1 & 86.8 & 58.5 & 72.6 \\
\midrule
\rowcolor[HTML]{DAE8FC} \multicolumn{11}{l}{\selfcall} \\
\rowcolor[HTML]{DAE8FC} \textbf{SubagentVL}${}^{\dagger}$ (Ours) & 7B & 93.0 & 89.5 & 91.6 & 93.3 & 60.8 & 77.0 & 87.0 & 58.3 & 72.6 \\
\rowcolor[HTML]{DAE8FC} $\Delta$ (vs Qwen2.5-VL-7B) &  & +19.1 & +22.4 & +20.4 & +8.1 & +8.6 & +8.2 & +8.2 & +6.5 & +7.3 \\
\rowcolor[HTML]{DAE8FC} $\Delta$ (vs DeepEyes) &  & +1.7 & +1.3 & +1.2 & +1.8 & +1.9 & +1.9 & +0.2 & -0.3 & 0.0 \\
\bottomrule
\end{tabular}
\label{tab:my-table}
\end{table*}

\paragraph{Training with \modelname{} is more efficient.}

The {$\vstar$ Benchmark} and {HR-Bench} assess reasoning over ultra-high-resolution images (2K–8K), where regions of interest are typically confined within 200 pixels. This extreme scale disparity poses challenges for base VLMs, which often fail to localize and reason correctly without auxiliary visual cues. Without relying on SFT initialization or in-context exemplars, both \textbf{DeepEyes} and our \textbf{SubagentVL} achieve strong end-to-end reinforcement learning performance, outperforming open-source models and GPT-4o across all three benchmarks. 
Remarkably, despite operating under more constrained training resources—smaller batch size (256 vs. 56), fewer rollouts, and less data access—\textbf{SubagentVL} surpasses baselines on $\vstar$ and HR-Bench 4K, while remaining competitive on 8K.
These results highlight that training with the purely textual \modelname{} framework is more resource-efficient and scalable than image-intensive methods such as iMCoT.

\paragraph{Impact on General Visual Ability.}
To investigate how reinforcement learning influences the model’s general visual capability, we evaluate whether basic skills such as OCR and grounding exhibit any improvement. The model is examined off the reasoning framework using standard benchmarks including RefCOCO\cite{kazemzadeh-etal-2014-referitgame}, OCR~\cite{liu2024ocrbench}, and POPE~\cite{Li-hallucination-2023}. Compared with DeepEyes, the results show that {SubagentVL} demonstrates only marginal gains in grounding and OCR~(\cref{tab:ocr,tab:visual-grounding}), while achieving a notable improvement in hallucination benchmarks (\cref{tab:pope}). This outcome is reasonable, since only the reasoning trajectories within the main agent are optimized during RL, while the interleaved subagent responses are masked out from log-probability computation. In contrast to the substantial performance boost observed in complex visual reasoning tasks with \modelname{}, this suggests that the enhanced ability primarily emerges from learning an appropriate subagent-calling strategy rather than from improvements in low-level visual perception.


\begin{table}
\centering
\small
\caption{Visual Grounding Benchmark Results. * indicates reproduced results.}
\label{tab:visual-grounding}
\begin{tabular}{l|ccc}
\toprule
Model & \multicolumn{1}{l}{refCOCO} & \multicolumn{1}{l}{refCOCO+} & \multicolumn{1}{l}{refCOCOg} \\
 \midrule
Qwen2.5-VL-7B* & 88.19 & 81.62 & 82.86 \\
DeepEyes* & 88.95 & 82.56 & 83.5 \\
SubagentVL (Ours) & 88.56 & 81.97 & 82.96 \\
\bottomrule
\end{tabular}
\end{table}

\begin{table}
\centering
\small
\caption{OCR Benchmark Results.}
\label{tab:ocr}
\begin{tabular}{l|ccc}
\toprule
 & {Qwen2.5-VL} & DeepEyes & SubagentVL \\
 \midrule
  OCRBench &0.844 & 0.846 & 0.845 \\
\bottomrule
\end{tabular}
\end{table}

\begin{table}
\small
\centering
\caption{POPE Benchmark Results for Hallucination Evaluation. Higher is better. * indicates reproduced results.}
\label{tab:pope}
\begin{tabular}{l|cccc}
\toprule
Model & \multicolumn{1}{l}{Adv.} & \multicolumn{1}{l}{Pop.} & \multicolumn{1}{l}{Rand.} & \multicolumn{1}{l}{Overall} \\
 \midrule
QwenVL-2.5-7B & 85.9 & 86.5 & 87.2 & 85.9 \\
DeepEyes & 84.0 & 87.5 & 91.8 & 87.7 \\
\midrule
QwenVL-2.5-7B* & {86.6} & {87.7} & {88.7} & 87.7 \\
SubagentVL (Ours) & 87.2 & 88.5 & 89.7 & 88.4 \\
\bottomrule
\end{tabular}
\end{table}

\begin{figure}
    \centering
    \includegraphics[width=\linewidth]{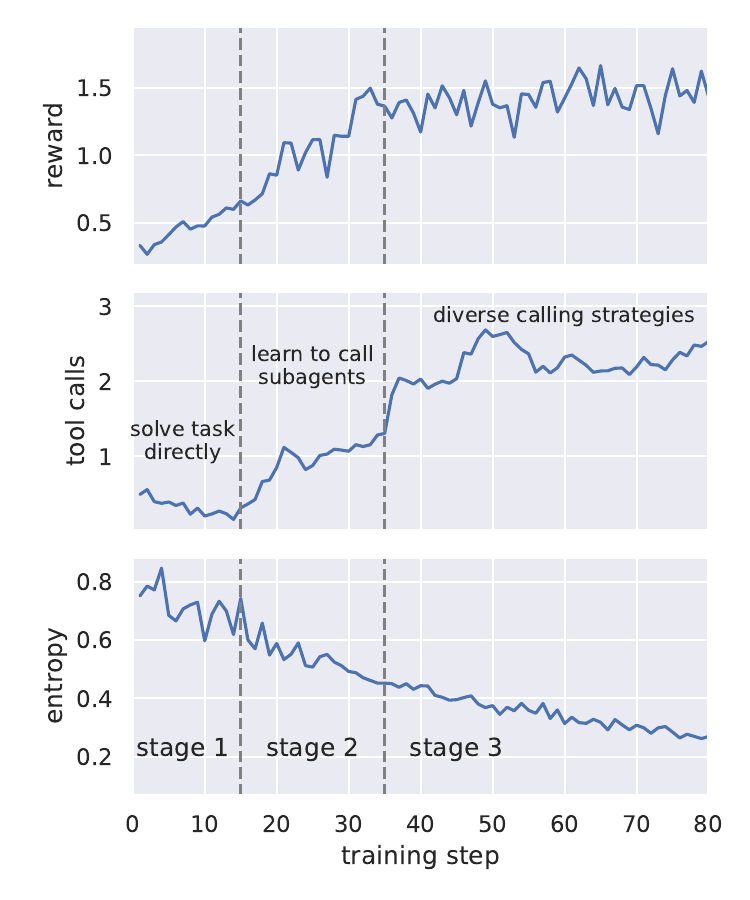}
    \caption{Training Dynamics. Training process could also be divided into three stages.}
    \label{fig:training-dynamics}
\end{figure}

\paragraph{Training Dynamics.}
We investigate how the model’s tool-calling behavior evolves throughout the training process. By jointly analyzing the number of tool calls and the corresponding reward trends in \cref{fig:training-dynamics}, we identify three distinct stages. In \textbf{the first stage}, the frequency of tool calls drops sharply while the reward gradually increases, accompanied by large fluctuations in entropy, contrary to intuition. Later ablation results indicate that strict tool-calling constraints on the three parameters (task type, prompt, bounding) limit the main agent’s capability to formulate subtasks.
This suggests that the main agent initially prefers to solve problems independently rather than invoking subagents. During \textbf{the second stage}, the number of tool calls begins to rise, leading to a rapid improvement in reward as the agent learns to effectively delegate complex subtasks to subagents. Notably, the transition between the first two stages does not occur at a fixed training step.  Finally, \textbf{in the third stage}, the agent consistently issues more subagent calls, indicating a matured coordination strategy between the main agent and subagents for handling challenging reasoning tasks.

\subsection{Ablation Study}

\paragraph{Constraints on Tool Calling Protocol.} 
When designing the tool-calling format in \cref{sec:tool-calling-protocol}, an important question arises regarding whether arguments should be strictly constrained. Specifically, we examine whether string-type arguments (\textit{task type} and \textit{prompt}) can be left empty, and whether the bounding-box argument should be optional—where an empty bounding box assigns the entire image to the subagent.

\begin{figure}
    \centering
    \includegraphics[width=0.9\linewidth]{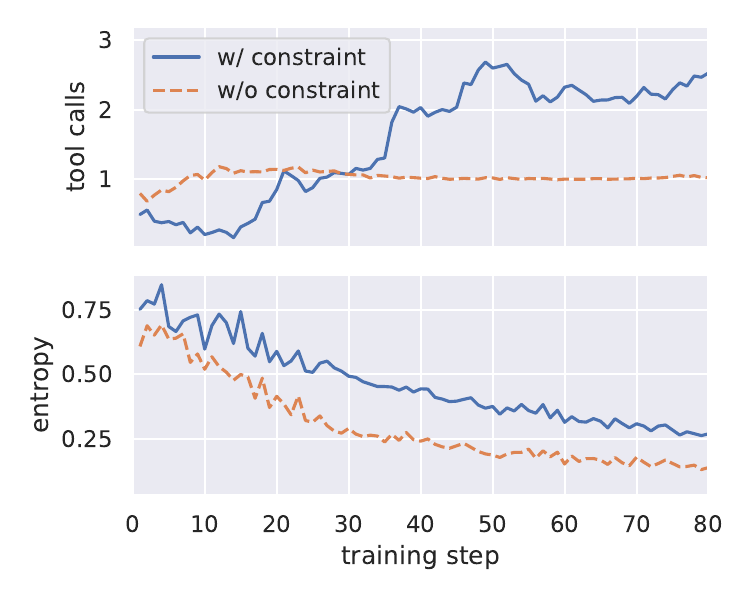}
    \caption{Training dynamics \wrt existence of constraints on tool calling protocol.}
    \label{fig:ablation-protocol}
\end{figure}

As shown in \cref{fig:ablation-protocol}, relaxing these constraints leads to undesirable learning dynamics. 
Although number of tool calls no longer decreases in the early stage of training, it quickly converges to one, indicating the absence of complex calling strategies. In such cases, the final calling pattern typically degenerates into simply repeating the user query while cropping a single object region, resulting in poor reasoning diversity. Quantitatively, as shown in \cref{tab:ablation-protocol}, the unconstrained variant achieves significantly lower than the constrained version, suggesting that enforcing stricter argument requirements encourages the agent to explore more structured and effective tool-calling behaviors.

\begin{table}[]
\centering
\caption{Evaluation on $V^*$ \wrt tool calling protocol constraints.}
\label{tab:ablation-protocol}
\begin{tabular}{ccccc}
\toprule
 constraints & {avg. calls} & {direct} & {relative} & {overall} \\
\midrule
w/o  & 0.8 & 89.6 & 80.3 & 85.9 \\
w/ & 1.5 & 93.0 & 89.5 & 91.6 \\
\bottomrule
\end{tabular}
\end{table}

\paragraph{Reward Design.}  
In the RL training setup of DeepEyes, the reward function does not include the ordering indicator \(\mathbb{I}_{\mathrm{tool} \prec \mathrm{ans}}\). In contrast, this term proves essential during the training of \subagent{}. As tool calls decrease sharply in the early training stage, \subagent{} tends to solve problems independently rather than invoking subagents. Without \(\mathbb{I}_{\mathrm{tool} \prec \mathrm{ans}}\), the model often abuses the reward function by repeatedly appending tool calls after the \texttt{</answer>} tag, leading to reward hacking and unstable policy learning. Incorporating this ordering constraint effectively prevents such behavior and promotes proper sequencing of tool use prior to final answer generation.

\begin{figure}
    \centering
    \includegraphics[width=0.9\linewidth]{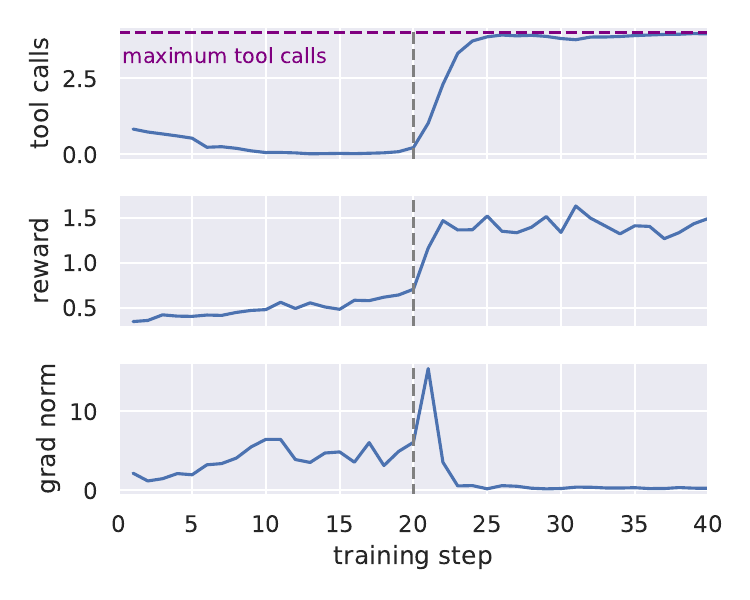}
    \caption{Reward hacking occurs at step 20 when the ordering constraint is missing.}
    \label{fig:reward-hacking}
\end{figure}

\paragraph{Data.}

\begin{figure}
    \centering
    \includegraphics[width=0.9\linewidth]{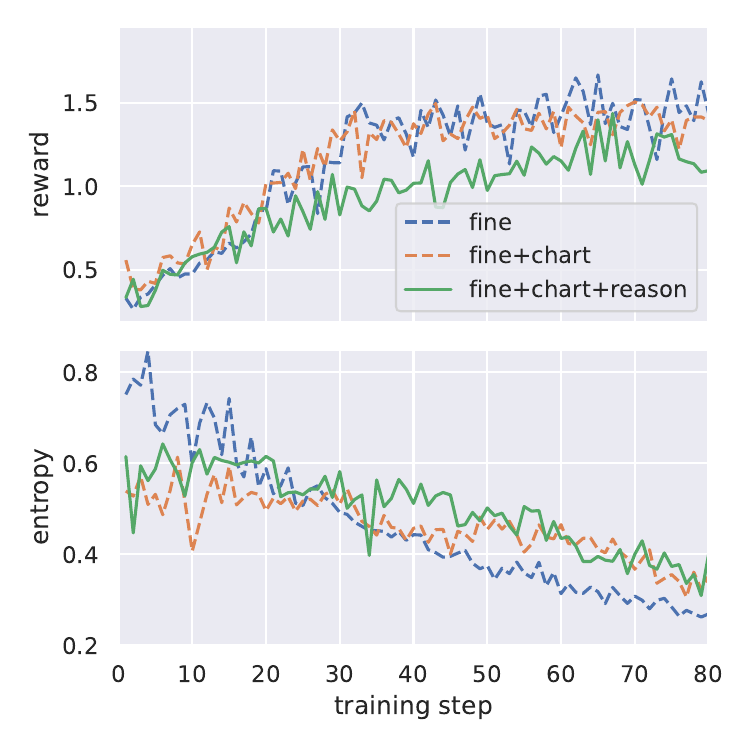}
    \caption{Ablation study on impact of training data.}
    \label{fig:ablation-dataset.}
\end{figure}

We further investigate how different data types affect the learning dynamics of \subagent{}. 
Recall that our training corpus consists of three categories: \textit{Fine} data (high-resolution or detail-focused images), \textit{Chart} data (complex structured plots), and \textit{Reason} data (smaller images paired with arithmetic reasoning, commonsense inference, or problem solving). 
As shown in \cref{fig:ablation-dataset.}, incorporating more diverse data (Fine + Chart) stabilizes the training process: the entropy decreases more gradually, and the model maintains scores on HR-Benchmarks.
However, adding \textit{Reason} data leads to degraded performance. Both the training reward curves and final evaluation metrics show clear drops when Reason data is included. 
We attribute this to a mismatch between the nature of \textit{Reason} dataset and the reasoning trajectories of \modelname{}. 
While Fine and Chart data encourage \modelname{} to attend to small visual regions, textual details, and spatially localized structures, the Reason data emphasizes abstract symbolic reasoning rather than visual grounding ability. 
This shifts the model’s focus away from precise region-based perception and tool-calling strategies, ultimately hindering its ability to generalize on high-resolution visual tasks. 
These findings highlight that, for \modelname{}, data that strengthens fine-grained visual discrimination contributes more effectively than data emphasizing non-visual reasoning.

\begin{table}[]
\small
\centering
\caption{Ablation study on impact of training data.}
\label{tab:data-impact}
\begin{tabular}{l|ccc|cc}
\toprule
 & Fine & Chart & Reason & HR4K & HR8K \\
 \midrule
Qwen2.5-VL &  &  &  & 68.8 & 65.3 \\
\midrule
\multirow{3}{*}{SubagentVL} & \checkmark &  &  & 77.0 & 72.6 \\
 & \checkmark & \checkmark &  & 76.3 & 72.6 \\
 & \checkmark & \checkmark & \checkmark & 74.4 & 70.3 \\
 \bottomrule
\end{tabular}
\end{table}
\section{Conclusion}
We proposed the \textit{\textbf{Self-Calling Chain-of-Thought}}~(\modelname{}), a new paradigm that reformulates interleaved multimodal CoT reasoning trajectories into a sequence of atomic subtasks coordinated by the main self-calling agent. 
This design removed the reasoning burden of managing multimodal CoTs and enables efficient end-to-end RL.
Despite limited computational resources, \subagent{} achieves striking performance on $\vstar$ and HR Bench, outperforming open-source models and surpassing prior multimodal Chain-of-Thought  paradigms and manually defined workflows. 
Ablation studies further show that strict tool-calling protocols, proper reward ordering constraints, and carefully curated training data are essential for stable learning and effective delegation. 
There findings highlight that structured textual decomposition, combined with lightweight visual subtasks, offers a scalable and resource-efficient solution for high-resolution multimodal reasoning.
{
    \small
    \bibliographystyle{ieeenat_fullname} 
    \bibliography{main}
}


\end{document}